\documentclass{article}
\usepackage{spconf,amsmath,graphicx,hyperref}
\usepackage{times}
\usepackage{epsfig}
\usepackage{amssymb}
\usepackage{enumerate}
\usepackage{multirow}
\usepackage{booktabs}
\usepackage{tabularx}
\usepackage{enumitem}

\usepackage{color}
\usepackage{xcolor}

\title{TP-MVCC: Tri-plane Multi-view Fusion Model for Silkie Chicken Counting}

\name{
Sirui Chen\qquad Yuhong Feng\qquad Yifeng Wang\qquad Jianghai Liao\qquad 
Qi Zhang\sthanks{Corresponding author.}
}
\address{College of Computer Science and Software Engineering, Shenzhen University}
\begin{document}
%
\maketitle
\begin{abstract}
Accurate animal counting is essential for smart farming but remains difficult in crowded scenes due to occlusions and limited camera views.  To address this, we propose a tri-plane-based multi-view chicken counting model (TP-MVCC), which leverages geometric projection and tri-plane fusion to integrate features from multiple cameras onto a unified ground plane. The framework extracts single-view features, aligns them via spatial transformation, and decodes a scene-level density map for precise chicken counting. In addition, we construct the first multi-view dataset of silkie chickens under real farming conditions. Experiments show that TP-MVCC significantly outperforms single-view and conventional fusion comparisons, achieving 95.1\% accuracy and strong robustness in dense, occluded scenarios, demonstrating its practical potential for intelligent agriculture.
\end{abstract}
\section{Introduction}

Animal perception, such as detection and counting, plays a critical role in intelligent agriculture and modern farm management. However, accurately estimating animal numbers remains highly challenging in real-world environments due to three major factors: (i) the natural crowding of animals that leads to severe occlusions in the scene, (ii) the narrow field-of-view (FoV) of individual cameras, and (iii) the frequent free movement of animals across large areas. As a result, conventional single-view counting or detection systems often exhibit significant accuracy degradation in dense/occluded scenarios \cite{4270159,7410729,mcnn,8578218,CCTrans}.

To address these limitations, multi-view counting has emerged as a promising solution. By deploying multiple overlapping cameras at different viewpoints, multi-view approaches can mitigate blind spots and occlusions while providing complementary scene information. Recent deep learning–based methods have leveraged this paradigm to improve robustness and accuracy in crowd counting, for example, by projecting features from multiple views onto a unified ground plane representation \cite{8953461,9577502} or by employing transformer architectures to encode camera parameters and fuse scene-level volumetric features \cite{CountFormer2024}. Nevertheless, existing methods are primarily developed for human crowds, while animal counting introduces unique challenges such as smaller individual size, greater movement freedom, and higher density under constrained environments, which exacerbate occlusion and viewpoint inconsistency.

Animal counting studies in the literature mainly rely on object detection frameworks such as YOLO and its variants \cite{agriculture12101659,ani14081227,LI2022107347} or regression-based density estimation approaches \cite{CAMPBELL2024108591,XU2020105300,app15073840}. These methods achieve good performance in low- and medium-density scenarios, but face severe bottlenecks in high-density poultry environments, where occlusions and missed detections lead to unsatisfactory accuracy. In addition, most existing works are restricted to single-view settings, which fundamentally limits their scalability to wide-area or large-scale farms.

\begin{figure*}[t]  
    \centering     
    \includegraphics[width=\textwidth]{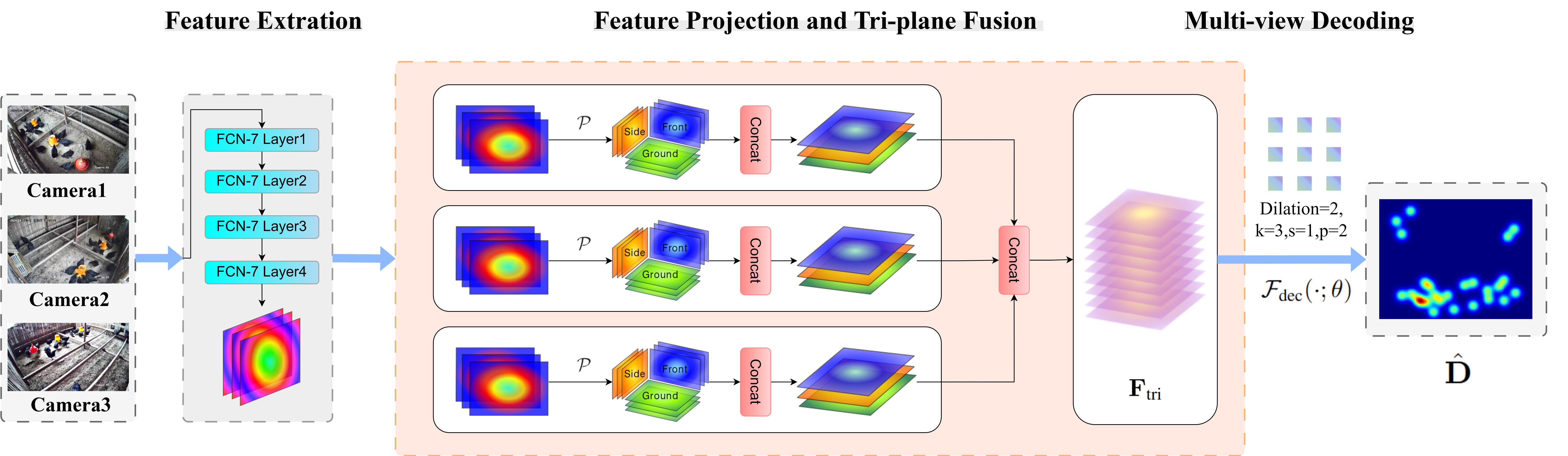}  
    \vspace{-0.6cm}
    \caption{The pipeline of the tri-plane-based multi-view fusion silkie chicken counting model (TP-MVCC).
    }
    \label{fig:pipeline}
     \vspace{-0.3cm}
\end{figure*}

In this paper, we propose a tri-plane-based multi-view chicken counting model (TP-MVCC) that integrates geometric projection and deep feature fusion for accurate counting in dense, occluded farming scenarios. The method first extracts view-specific features using a fully convolutional backbone, then projects and fuses them onto a unified ground plane via a tri-plane multi-view feature fusion strategy, and finally decodes the fused scene representation into a ground-plane density map for precise counting. To support this study, we construct the first multi-view silkie chicken dataset captured under real farming conditions, providing diverse scenes for evaluation. The main contributions of this work are threefold:
\begin{itemize}
    \item We present an end-to-end trainable DNN-based multi-view silkie chicken counting model (TP-MVCC) that fuses information from multiple camera views to generate a ground-plane density map for a chicken coop scene. To the best of our knowledge, this is the first study to explore multi-view density map estimation for animal counting.
    \item We have collected a unique dataset of silkie chickens from multiple camera views. To the best of our knowledge, this is the first dataset of its kind specifically focused on silkie chickens.
    \item We compare the proposed TP-MVCC method with other single-view and multi-view approaches, demonstrating its advantage for silkie chicken counting.
\end{itemize}

\section{TP-MVCC MODEL}

The proposed tri-plane-based multi-view fusion chicken counting model (TP-MVCC) aims to estimate scene-level ground-plane density maps from synchronized multi-view inputs. The design relies on two assumptions: (1) cameras remain fixed after calibration, with accurate intrinsic and extrinsic parameters; (2) temporal synchronization ensures spatial consistency among views. As in Fig.~\ref{fig:pipeline}, TP-MVCC consists of three stages: First, discriminative multi-scale features are extracted from each view using a lightweight fully convolutional network (FCN-7); Second, these features are geometrically aligned via a spatial transformer and projected onto three planes (ground, front, and side) to enable cross-view correspondence;  Finally, the tri-plane features are fused and decoded into the scene-level density map.

\textbf{Single-view Feature Extraction:} Each camera view is processed by a lightweight fully convolutional network (FCN-7) to extract high-level multi-scale features, preserving both local details (e.g., feather texture, limb contour) and global distribution patterns. FCN-7 consists of 7 convolutional layers with dilated convolutions to balance feature expressiveness and computational efficiency. This module provides discriminative representations for subsequent multi-view fusion.

\textbf{Feature Projection and Tri-plane Fusion:} Using camera intrinsic and extrinsic parameters, each view’s feature map is projected onto a unified 3D ground plane. Since the exact height of each pixel is unknown, we assume a fixed center height for each chicken. The projection is implemented via a differentiable spatial transformation:

\begin{equation}
	\mathbf{F}_{\text{ground}}(p) = \sum_{q \in \mathcal{N}(p)} \mathbf{F}_{\text{image}}(q) \cdot w(p,q),
	\label{eq:bilinear_sampling}
\end{equation}
Where $p$ is the target grid point coordinates, $q$ is the source image domain coordinates, and the interpolation weight $w(p,q)$is defined as follows.
\begin{equation}
	w(p,q) = \max(0, 1 - |x_q - x_p|) \cdot \max(0, 1 - |y_q - y_p|).
	\label{eq:bilinear_weight}
\end{equation}
By definition, $\mathcal{N}(p)=\{(i_0,j_0),(i_0,j_1),(i_1,j_0),(i_1,j_1)\}$ corresponds to the four integer neighbors of the projected location $(\tilde{u},\tilde{v})$ in the source image domain.

\begin{figure}[t]
    \centering
    \includegraphics[width=0.9\columnwidth]{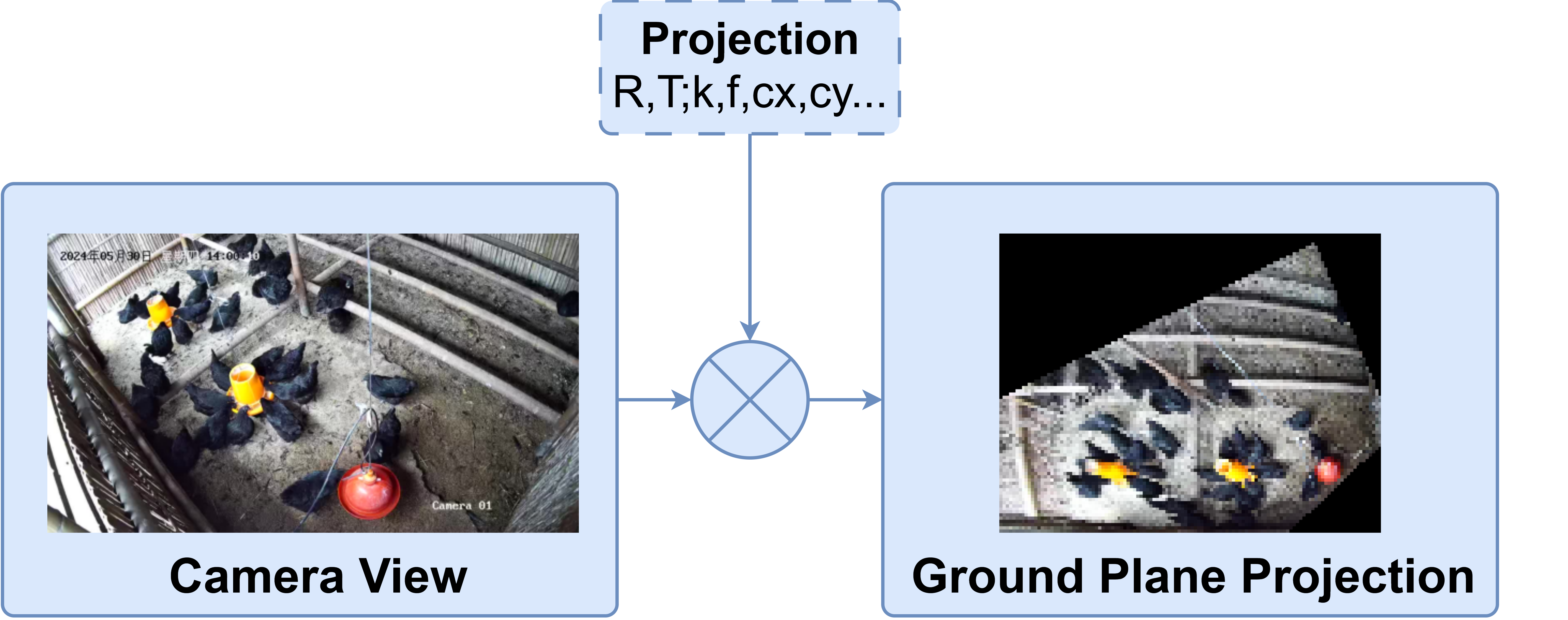}
    \vspace{-0.4cm}
    \caption{Projection module based on spatial transformation.
    }
    \vspace{-0.5cm}
    \label{fig:sample}
\end{figure}

To enhance multi-view alignment, projected features are divided into three planes—ground, front, and side—capturing complementary spatial information. Features from all views are geometrically aligned and fused on each plane, reducing perspective-induced discrepancies and improving cross-view correspondence. This tri-plane fusion enables effective integration of multi-view information, resulting in more accurate and robust density estimation. Fig.  \ref{fig:sample} illustrates the projection and spatial transformation process.

\textbf{Multi-view Decoding Network: } 
After tri-plane feature fusion, the combined feature map from all projection planes, denoted as $\mathbf{F}_{\text{tri}} \in \mathbb{R}^{C \times H \times W}$, is concatenated along the channel dimension to enrich spatial representation. 
A lightweight decoder $\mathcal{F}_{\text{dec}}(\cdot;\theta)$ then regresses the final scene-level density map:
\begin{equation}
    \hat{\mathbf{D}} = \mathcal{F}_{\text{dec}}(\mathbf{F}_{\text{tri}}; \theta)
    \label{eq:decode_density}
\end{equation}

Supervision is provided by the ground-truth density maps $\mathbf{D}$, using pixel-wise mean squared error (MSE) loss:
\begin{equation}
    \mathcal{L}_{\text{MSE}} = \frac{1}{N}\sum_{i=1}^{N}\|\hat{\mathbf{D}}_i - \mathbf{D}_i\|_2^2,
    \label{eq:mse_loss}
\end{equation}
where $N$ is the total number of pixels. This formulation ensures accurate regression of density distributions while preserving the total scene-level count through integration over the ground plane.


\textbf{Model Training:}
The model training is as follows.

\begin{itemize}[noitemsep, topsep=1pt]
    \item \textbf{Stage 1 (Single-view pretraining):} 
    The FCN-7 backbone is trained using all single-view images with their corresponding density maps as supervision. This stage learns a unified feature extractor capable of capturing discriminative per-view features across all camera views via pixel-wise MSE loss. 

    \item \textbf{Stage 2 (Fusion training):} 
    Only scene-level density maps are used as supervision. The fusion and decoding modules are trained while the single-view backbone is frozen, enabling the network to learn multi-view feature integration. 

    \item \textbf{Stage 3 (End-to-end fine-tuning):} 
    All network modules are jointly updated to further refine feature extraction, multi-view fusion, and density map prediction. Pixel-wise MSE loss is used, with a small batch size and a gradually decaying learning rate for stable convergence.
\end{itemize}

\section{Experiments}

\begin{figure}[t]
    \centering
    \includegraphics[width=\columnwidth]{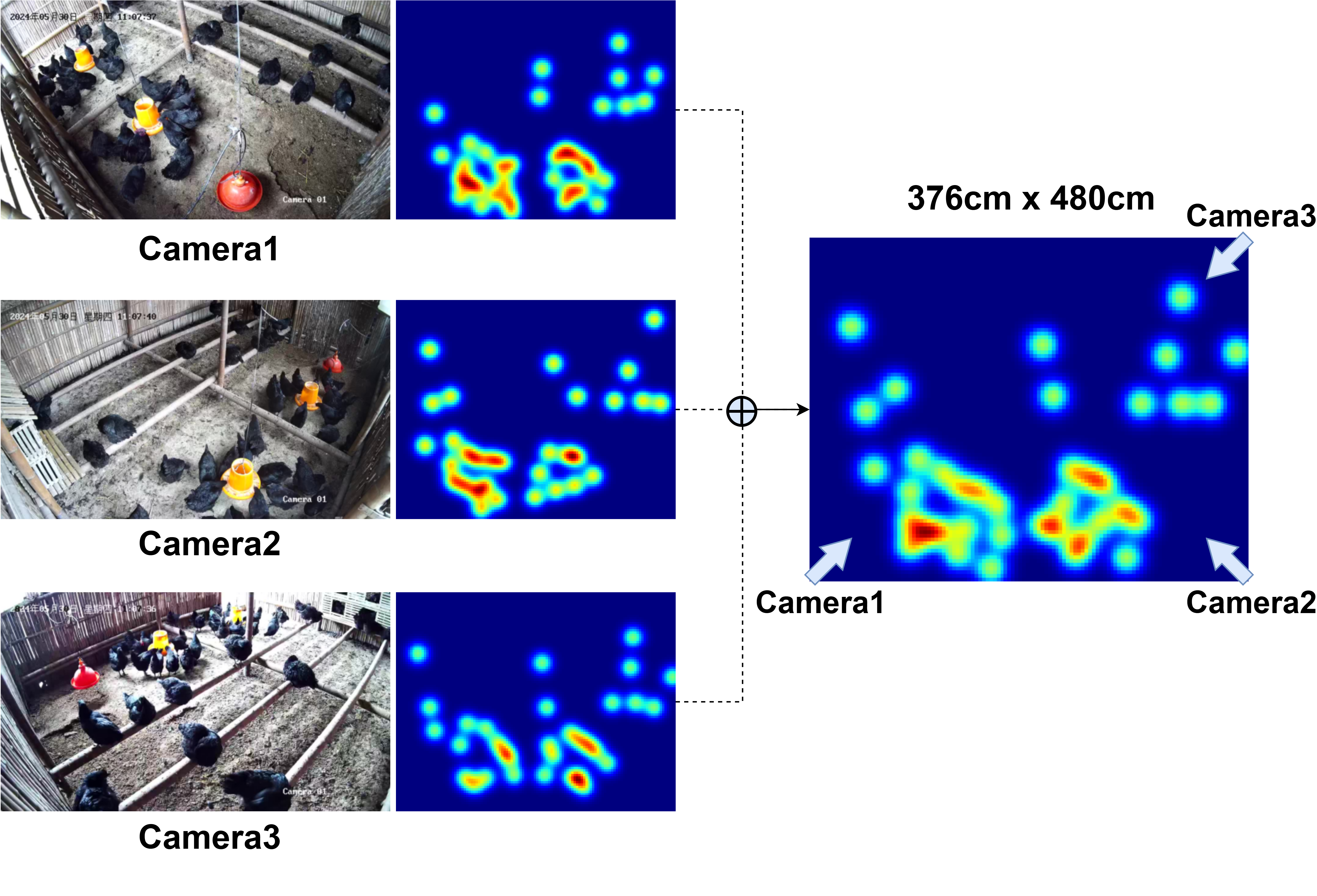}
    \vspace{-0.7cm}
    \caption{Example of the proposed dataset with three synchronized viewpoints. 
    Cameras were mounted at a tilt angle of $\sim$30° to cover different regions of the chicken shed.
    }
    \label{fig:dataset}
    \vspace{-0.5cm}
\end{figure}

\subsection{Dataset}
To the best of our knowledge, this is the first multi-view dataset dedicated to silkie chicken counting in real farming conditions. It covers diverse behavioral scenarios, including feeding, roosting, and sleeping, with synchronized images captured from three fixed cameras (Fig.~\ref{fig:dataset}). To handle height variations between ground and railing, a dual-plane projection strategy is adopted, with an IP-based matching algorithm fusing projected density maps into a unified ground coordinate system. Ground-truth annotations are obtained via semi-automatic SAM \cite{sam} segmentation on ISAT, followed by camera calibration. 

The dataset consists of 60,000 frames recorded at 10 fps, from which 400 frames are uniformly sampled and annotated. It is split into 200 training frames (avg.~36.84 chickens/frame) and 200 validation/test frames (avg.~38.13 chickens/frame). The selected frames span multiple behavioral states, primarily feeding and roosting. While limited to a single controlled coop and the Silkie breed, this dataset establishes a benchmark for future research on multi-view animal counting.

\subsection{Experiment settings}


\textbf{Comparisons.} 
We evaluate TP-MVCC against five representative methods.  constructed from two single-view feature extractors and two multi-view fusion strategies, along with a detection-based method. 
Specifically, \textit{MAN}~\cite{MAN} and \textit{CSRNet}~\cite{8578218} serve as density-based single-view backbones. 
For multi-view fusion, we consider \textit{Mask Fusion(MF)}~\cite{7485869}, which restricts counting to masked near-field regions and fuses results via geometric priors, thereby reducing the influence of irrelevant areas, and \textit{Density-map Weighted Fusion (DWF)}~\cite{RYAN201498}, which merges single-view density maps through pre-defined perspective-aware weights.
By combining the two extractors with the two fusion strategies, we obtain four baselines: MAN+Mask, MAN+DWF, CSRNet+Mask, and CSRNet+DWF. 
In addition, we include \textit{YOLOv11s}~\cite{YOLOv11}, a state-of-the-art detector adapted for chicken counting by detecting individual instances per view. 
Together, these baselines span both regression- and detection-based paradigms, as well as early and late fusion strategies, providing a comprehensive benchmark for comparison with our method.

\textbf{Metrics.} 
We report Mean Squared Error (MSE), Mean Absolute Error (MAE), and frame-level accuracy (\textit{Rate}). True counts 
are the individual numbers covered by all camera views, which are obtained by integrating over the scene-level density maps.


\begin{table}[t]
  \centering
  \caption{Comparison on the silkie chicken dataset. GPU Memory usage is measured in MB. }
  \setlength{\tabcolsep}{4pt} 
  \begin{tabular}{l|cccc}
    \toprule
    Method & MSE$\!\downarrow$ & MAE$\!\downarrow$ & Rate$\!\uparrow$ & GPU Mem$\!\downarrow$ \\
    \midrule
    MAN\cite{MAN}+MF\cite{7485869}   & 8.01  & 6.68  & 0.840 & 1919.16 \\
    MAN\cite{MAN}+DWF\cite{RYAN201498}    & 9.54 & 6.99  & 0.828 & 1919.16 \\
    CSRNet\cite{8578218}+MF\cite{7485869}    & 9.67   & 7.13   & 0.831   & 3418.90 \\
    CSRNet\cite{8578218}+DWF\cite{RYAN201498}     & 11.87   & 8.48   & 0.813   & 3418.90 \\
    YOLO11s\cite{YOLOv11}        & 8.09 & 7.52 & 0.842 & 408.96 \\
    \midrule
    \textbf{TP-MVCC (ours)} 
                   & \textbf{3.84} & \textbf{2.89} & \textbf{0.951} & 1648.47 \\
    \bottomrule
  \end{tabular}
  \label{tab:res}
  \vspace{-0.5cm}
\end{table}

\subsection{Experiment results}
\begin{figure}[t]
    \centering
    \includegraphics[width=1.0\columnwidth]{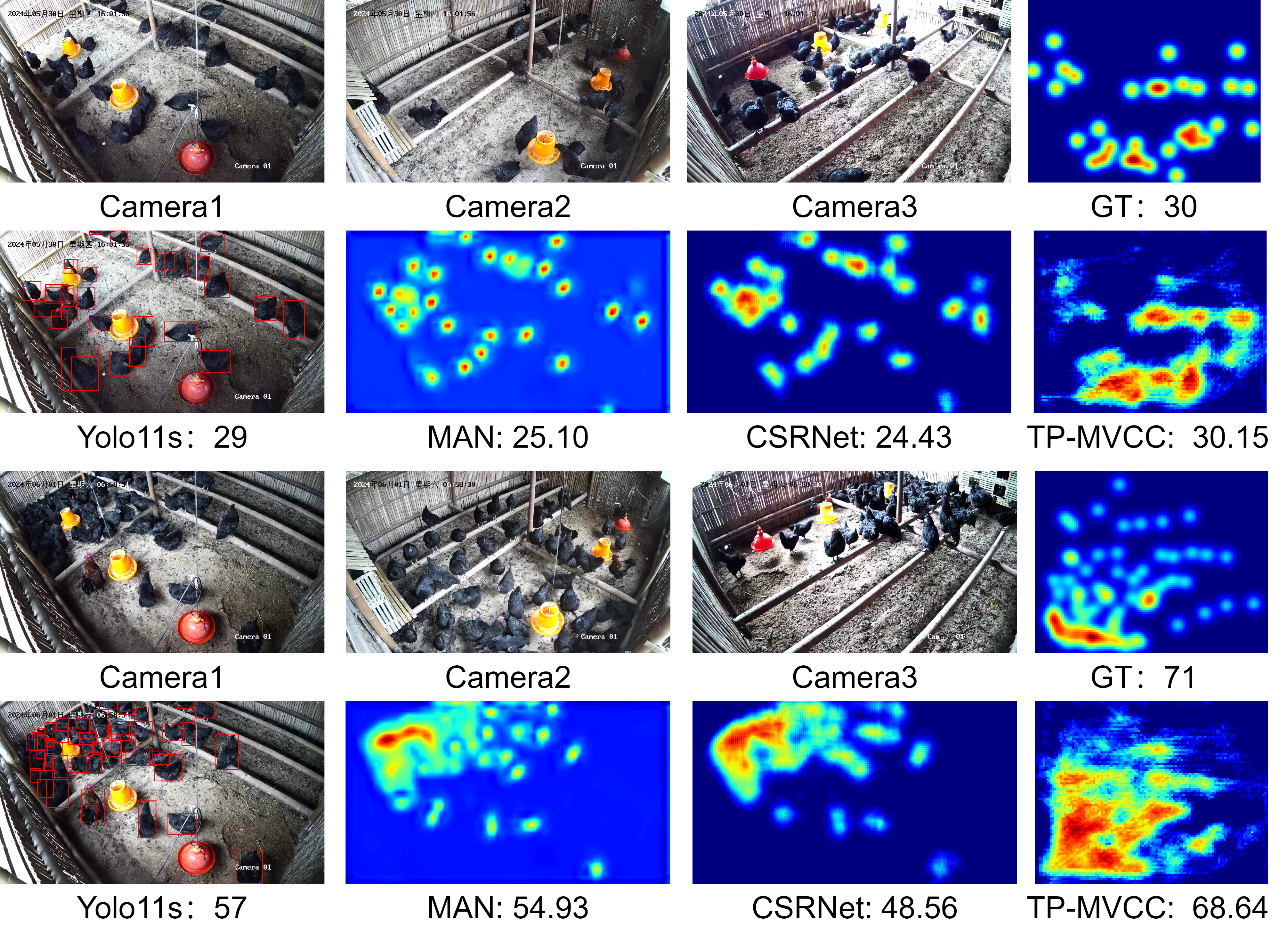}
    \vspace{-0.5cm}
    \caption{Qualitative results on two examples. 
    For each example, the \textbf{first row} shows the three input views (Camera~1–3) together with the ground-truth ground-plane density map. 
    The \textbf{second row} presents the predictions of comparison methods on Camera~1: YOLOv11s, MAN, and CSRNet, as well as the result of our TP-MVCC model on the ground plane. 
    }
    \label{fig:res}
    \vspace{-0.5cm}
\end{figure}

As shown in Table~\ref{tab:res}, our TP-MVCC consistently outperforms all competitors by a large margin, achieving the lowest error (MSE=3.84, MAE=2.89) and the highest accuracy (Rate=0.951).  
Among the baselines, YOLO11s benefits from instance-level detection but struggles under heavy occlusion, while CSRNet and MAN rely on single-view density estimation, making them sensitive to perspective distortion and limited coverage. Multi-view fusion strategies (Mask and DWF) improve robustness but still suffer from error accumulation due to per-view predictions.  In contrast, our end-to-end multi-view fusion captures complementary information across cameras and models the ground-plane distribution, yielding superior performance (see Fig.~\ref{fig:res} for qualitative comparisons).

Regarding resource usage (Table~\ref{tab:res}), we acknowledge that TP-MVCC requires more memory than YOLO11s (1648.47 MB vs. 408.96 MB), while remaining substantially more efficient than CSRNet and MAN, and achieving the best accuracy. In practical deployment, the system samples one frame per hour from synchronized streams and processes three views efficiently on an NVIDIA RTX 3090, demonstrating both accuracy and suitability for smart farming.

\begin{table}[t]
  \centering
  \caption{Ablation on training strategy. We compare two settings for the single-view feature extractor: freezing the parameters after pre-training vs. allowing them to be updated during end-to-end training.  }
    \begin{tabular}{l|ccc}
    \toprule
    Parameters & $MSE\!\downarrow$ & $MAE\!\downarrow$ & $Rate\!\uparrow$\\
    \midrule
    Fixed & 4.39 & 3.56 & 0.924 \\
    Trainable & \textbf{3.84} & \textbf{2.89} & \textbf{0.951} \\
    \bottomrule
  \end{tabular}
  \vspace{-0.3cm}
  \label{tab:parameters}
\end{table}

\subsection{Ablation Study}  

\textbf{Training strategy.}  
Table~\ref{tab:parameters} shows that updating the parameters of the single-view feature extractor reduces MSE from 4.39 to 3.84 and MAE from 3.56 to 2.89.  
This end-to-end optimization enhances adaptability to complex scenes (e.g., occlusion and illumination changes), allowing the model to better capture inter-view variations.  

\textbf{View combinations.}  
As shown in Table~\ref{tab:combination}, single-view inputs yield large errors due to occlusion and limited coverage.
In contrast, Camera1\&3 achieve the best two-view performance (MSE=4.37, MAE=3.31), benefiting from complementary viewpoints and reduced overlap.  
All three views together yield the best overall accuracy, confirming the effectiveness of multi-view fusion in capturing diverse scene information.  

\textbf{Dilated convolution.}  
Table~\ref{tab:dilation} demonstrates that introducing dilated convolution ($dilation=2$) enlarges the receptive field, improving spatial context modeling.  
This is particularly beneficial in dense or heavily occluded scenes, where global dependencies help distinguish clustered chickens.

\begin{table}[t]
  \centering
  \caption{Ablation on camera combinations. We evaluate the effect of using different subsets of cameras: single views, all two-view pairs, and the full three-view setting. }

    \begin{tabular}{l|ccc}
    \toprule
    Combinations & $MSE\!\downarrow$ & $MAE\!\downarrow$ & $Rate\!\uparrow$\\
    \midrule
    Camera1 & 12.34  & 9.55  & 0.774 \\
    Camera2 & 16.31  & 14.44  & 0.631 \\
    Camera3 & 12.53  & 10.70  & 0.724 \\
    Camera1 \& Camera2 & 5.24  & 4.39  & 0.883 \\
    Camera1 \& Camera3 & \underline{4.37} & \underline{3.31} & \underline{0.924} \\
    Camera2 \& Camera3 & 4.53  & 3.72  & 0.917 \\
    \midrule
    \textbf{All (ours)} & \textbf{3.84 } & \textbf{2.89 } & \textbf{0.951} \\
    \bottomrule
    \end{tabular}%
  \vspace{-0.5cm}

  \label{tab:combination}%
\end{table}%

\begin{table}[t]
  \centering
  \caption{Ablation on dilated convolution. We compare models with different dilation rates in the convolutional layers. 
  }

  \begin{tabular}{c|ccc}
    \toprule
    Dilation & $MSE\!\downarrow$ & $MAE\!\downarrow$ & $Rate\!\uparrow$\\
    \midrule
    1     & 4.14 & 3.16 & 0.938 \\
    2     & \textbf{3.84} & \textbf{2.89} & \textbf{0.951} \\
    3     & 4.78 & 3.30 & 0.917 \\
    \bottomrule
  \end{tabular}
    \vspace{-0.5cm}

  \label{tab:dilation}
\end{table}

\section{Discussion and Conclusion}
Aiming at the challenging problem of silkie chicken counting in agricultural scenes, this paper proposes a trip-plane-based multi-view fusion model, TP-MVCC.  By fusing features from multiple camera views, the scene-level ground density map is predicted for high-precision counting, and semantic information from different views is adaptively integrated using dynamic learnable weights.  It overcomes single-view occlusion and limited field of view, and introduces dilated convolution to enhance context modeling and optimize counting in dense areas.  Moreover, we propose the first multi-view silkie chicken dataset, containing occlusion, illumination changes, and dynamic chicken behaviors in complex breeding scenes.  Although this study assumes fixed camera parameters calculated by additional algorithms, environmental factors and device adjustments may shift perspectives, introducing projection errors.  Adapting the framework to mobile cameras with unknown parameters remains important future work.

\bibliographystyle{IEEEbib}
\bibliography{refs}

\end{document}